\documentclass[11pt]{article}

\usepackage[preprint]{acl}
\usepackage{amsmath}
\usepackage{times}
\usepackage{latexsym}
\usepackage{booktabs}
\usepackage[T1]{fontenc}

\usepackage[utf8]{inputenc}

\usepackage{microtype}

\usepackage{inconsolata}

\usepackage{graphicx}
\usepackage{amssymb}
\usepackage{amsmath}

\title{VDLM: Variable Diffusion LMs via Robust Latent-to-Text Rendering}

\author{Shuhui Qu \\
  Stanford University \\
  \texttt{shuhuiq@stanford.edu} \\
  }

\begin{document}
\maketitle

\begin{abstract}
Autoregressive language models decode left-to-right with irreversible commitments, limiting revision during multi-step reasoning. We propose \textbf{VDLM}, a modular variable diffusion language model that separates semantic planning from text rendering. VDLM applies LLaDA-style masked diffusion over semantic variable embeddings to enable iterative refinement in latent space, then post-trains the planner with trajectory-aware optimization using embedding-space rewards and values, avoiding text decoding inside the RL loop. To convert planned embeddings back to text, we use a \textbf{Vec2Text} renderer and introduce \textbf{embedding perturbations} to robustify decoding under planner noise. Across nine benchmarks spanning general reasoning, math, and code, VDLM is competitive in pre-training and yields substantial post-training improvements on long-form generation tasks, outperforming other baselines. These results highlight the effectiveness of embedding-space post-training and robust latent-to-text rendering for diffusion language modeling.
\end{abstract}

\section{Introduction}

Multi-step reasoning and long-form generation remain challenging regimes for modern language models, especially when solutions require revising intermediate decisions rather than extending a locally consistent prefix\citep{schmidt2019generalization, he2021exposure,huang2023large}. Autoregressive (AR) based language models currently dominate the modeling paradigm for large language models (LLMs)\cite{minaee2024large,prabhudesai2025diffusion}. AR based LLM decode left-to-right with irreversible commitments \citep{brown2020language,touvron2023llama,achiam2023gpt}. This \emph{sequential commitment} makes revision expensive: correcting an early mistake typically requires restarting, branching, or relying on external search procedures. In long-horizon reasoning (e.g., mathematics or code), small local errors can cascade, because later tokens must condition on previously committed prefix tokens\cite{arora2022exposure,he2021exposure,yao2023tree}.

Recent work has tried to mitigate these issues by encouraging explicit intermediate reasoning (e.g., Chain-of-Thought, Tree-of-Thought) and sampling-based exploration \citep{wei2022chain,wang2022self,yao2023tree}. While these methods improve success rates in many settings, they do not fundamentally change the decoding pattern: the model commits token-by-token, and exploration remains limited by the inability to \emph{revise} already produced text without re-generating\cite{yu2025discrete,xiong2025unveiling}.

Diffusion-style generation offers a complementary perspective: instead of committing left-to-right, a model iteratively refines a partially observed representation \citep{ho2020denoising,austin2021structured}. Diffusion language models such as \textsc{LLaDA} scale this idea by learning to recover randomly masked tokens and sampling via repeated (re-)masking and denoising \citep{nie2025large, bie2025llada2, wang2025revolutionizing}. However, token-level refinement is still costly when outputs are long: the refinement unit is the token, so the model repeatedly solves a high-dimensional discrete reconstruction problem. Moreover, semi-autoregressive / block diffusion variants reduce wall-clock latency but can degrade quality for long segments, since errors introduced within a block are harder to correct once the block is committed \citep{nie2025large, israel2506accelerating,wang2018semi,lu2025adablock}.

We address this tension by shifting diffusion \emph{above tokens}. Our key premise is that long-form generation should be driven by a \emph{short, structured planner} and a \emph{strong renderer} that faithfully reconstructs text from the plan. Concretely, we propose \textbf{VDLM}, a \emph{variable diffusion language model} that separates \textbf{semantic planning} from \textbf{text rendering}. VDLM operates on a sequence of continuous \emph{semantic variable embeddings}, which are constructed as question blocks, answer options, or reasoning concept segments, and refines these embeddings with masked denoising. This yields a much shorter refinement horizon than token diffusion while retaining the ability to revise and globally coordinate decisions.

Our framework consists of three components:
\begin{enumerate}
    \item \textbf{Variable diffusion planning.} We apply \textsc{LLaDA}-style masked diffusion to a matrix of variable embeddings, iteratively refining a latent plan $V$ in continuous space \citep{nie2025large}.
    \item \textbf{Embedding-space post-training.} To align the planner with downstream task metrics, we adapt \textsc{TraceRL} to latent variable space \citep{wang2025revolutionizing}. Both the reward signal and the value function are computed from variable embeddings. There is \emph{no text decoding inside the RL loop}, making post-training efficient even when rendering is expensive.
    \item \textbf{Robust rendering via noise injection \textsc{Vec2Text}.} We invert planned embeddings back to text using \textsc{Vec2Text} \citep{morris2023text}. Since planner outputs are noisier than clean encoder embeddings, we perturb embeddings during renderer training (via $L_2$ noise) and train the corrector to recover the exact target text, substantially improving robustness under planner-induced distribution shift.
\end{enumerate}

A central motivation of this paper is that \textbf{rendering quality is the bottleneck} for long outputs: improving the latent-to-text renderer yields substantially larger gains than semi-AR token-level rendering when segments become long. VDLM explicitly targets this bottleneck by pairing semantic-unit planning with a robustly trained renderer trained to withstand latent noise.

\paragraph{Results overview.}
Across nine benchmarks spanning general knowledge, reasoning, mathematics/science, and code, VDLM is competitive in the pre-training setting and improves markedly after post-training. Notably, SFT+RL post-training in embedding space yields strong long-form gains. These results support the thesis that semantic-unit diffusion planning can be effectively aligned without decoding in the RL loop, and that robust latent-to-text rendering is decisive for long-sequence performance.

The rest of the paper is organized as following: Section~\ref{sec:method} introduces VDLM, including variable construction, the LLaDA-style masked denoiser, TraceRL-style embedding-space post-training, and robust \textsc{Vec2Text} rendering. Section~\ref{sec:experiments} presents experimental results on nine benchmarks together with ablations and efficiency analysis. Section~\ref{sec:discussion} discusses implications, limitations, and future directions.

\section{Related Work}
\label{sec:related_work}

\subsection{Masked diffusion language models}
Diffusion-style sequence models replace left-to-right decoding with iterative denoising under a forward corruption process, enabling global refinement rather than prefix-locked commitments. Early discrete or masked formulations include D3PM-style discrete diffusion and masked-token denoising frameworks \citep{austin2021structured, peebles2023scalable}. For text generation, Diffusion-LM explores diffusion formulations for language modeling and controllable generation in token space \citep{li2022diffusion,cheng2025sdar,he2023diffusionbert}. More recently, masked diffusion LMs have improved scalability and sampling practicality. MDLM develops simple masked diffusion objectives for language with strong empirical performance \citep{sahoo2024simple,ye2025dream}. LLaDA scales masked diffusion for language and supports multiple decoding strategies (pure diffusion and block/semi-AR sampling) under a unified training objective \citep{nie2025large}. 

\subsection{Latent-space reasoning and semantic-unit generation}
A growing line of work argues that allocating computation in \emph{latent} representations can improve reasoning by delaying irreversible surface-form decisions\citep{hao2024training}. Continuous latent reasoning approaches train models to perform intermediate computation in a learned continuous space and decode to text only at the end \citep{zhu2025emergence}. Related directions study how latent structures can be induced or discovered to support reasoning behavior, such as latent skill discovery for chain-of-thought reasoning \citep{xu2023latent}.

Complementary to latent reasoning, recent work explores operating over higher-level semantic units (“concepts”) rather than tokens. A parallel work, Dynamic Large Concept Models explicitly model and generate sequences of large semantic concepts to reduce token-level burden and improve efficiency \citep{qu2025dynamic}. VDLM follows this spirit but focuses on \emph{diffusion-time semantic revision}: masked diffusion is applied to a fixed-length sequence of variable slots in continuous space, and long text is recovered by a separate renderer.

\subsection{Reinforcement learning for diffusion language models}
Post-training diffusion LMs with reinforcement learning is non-trivial because generation is a multi-step denoising trajectory; effective credit assignment should respect the \emph{inference trace}. \textsc{TraceRL} addresses this by introducing trajectory-aware objectives and diffusion-aware value modeling to stabilize RL for diffusion generation \citep{wang2025revolutionizing}. \citep{wang2025d2} proposes a principled RL framework for masked DLMs, deriving policy-gradient estimators that explicitly rely on sampling-trajectory likelihood computation 

We adapt TraceRL-style ideas to \emph{variable diffusion} and further reduce training cost by computing rewards and values directly in embedding space, avoiding text rendering inside the RL loop.

\subsection{Embedding inversion and Vec2Text rendering}
Embedding inversion studies how much information is preserved in text embedding models and how to recover text from dense vectors. \textsc{Vec2Text} shows that modern embeddings retain substantial recoverable signal and proposes a hypothesizer-corrector architecture that iteratively refines a
decoded hypothesis by comparing its embedding to the target \citep{morris2023text}. Vec2Text is commonly trained to invert strong embedders such as GTR-style encoders \citep{morris2023text, morris2023language}. In practice, however, inversion can be brittle under off-manifold or noisy embeddings. Recent results further suggest that transformer representations can be injective and exactly invertible, enabling provable input reconstruction from hidden activations~\citep{nikolaou2025language}. CALM thus improves robustness of language autoencoders by training the encoder–decoder with variational regularization and noise injection so that decoding remains stable under perturbations in the continuous latent space~\citep{shao2025continuous}.

\section{Method}
\label{sec:method}

\subsection{Overview}

\begin{figure}[t]
    \centering
    \includegraphics[width=\linewidth]{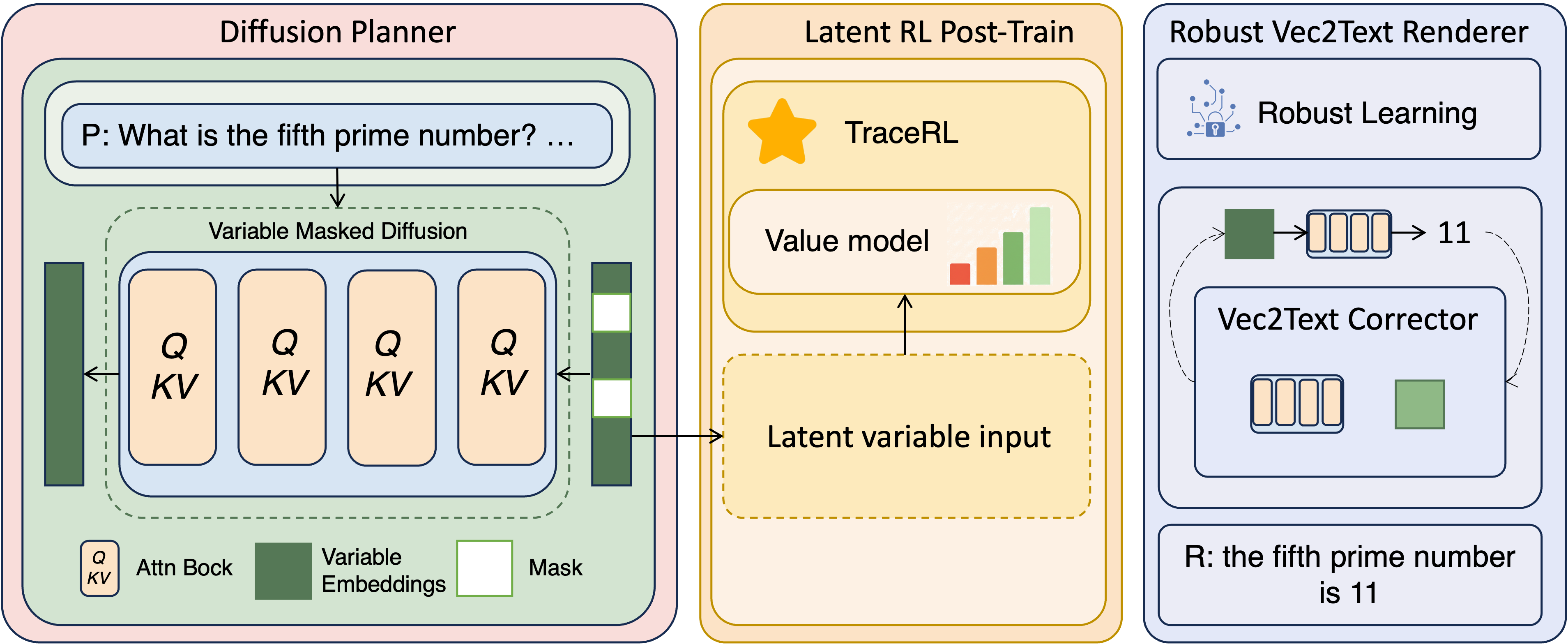}
    \caption{\textbf{VDLM pipeline overview.} A variable-level diffusion planner refines semantic variable embeddings via masked denoising; the planner is post-trained with TraceRL-style trajectory-aware RL entirely in latent space (reward/value computed from embeddings, no text decoding in the RL loop); an robustly trained Vec2Text renderer robustly converts the final latent plan into text under planner-induced embedding noise.}
    \label{fig:vdlm_pipeline}
\end{figure}

Given an input $p$, we construct a fixed-length sequence of semantic variables
$\tilde{V}=[\tilde{v}_1,\ldots,\tilde{v}_n]$ (e.g., prompt parts, options, or steps), pad to $N_{\max}$, and embed each variable into a continuous vector. We train a shared conditional denoiser to recover masked variables in embedding space (\textbf{Stage~I}) and then finetune it with TraceRL-style reward RL for downstream tasks (\textbf{Stage~II}) from the Stage~I checkpoint. Finally, a \textsc{Vec2Text} renderer decodes the planned embeddings to text (\textbf{Stage~III}); it is trained separately using paired text--embedding data from a frozen encoder $E(\cdot)$, with noise injected to ensure robustness. No gradients are propagated from the renderer back into the planner/encoder unless stated otherwise.

\subsection{Variable construction and embedding}
\label{sec:variables}

We obtain semantic variables by a lightweight segmentation function $\mathrm{Split}(\cdot)$ that produces
$n \le N_{\max}$ text spans:
\begin{equation}
\begin{split}
    \tilde{V}=\mathrm{Split}(p)&=\big[\tilde v_1,\ldots,\tilde v_n\big]\\ n&=|\tilde V|\le N_{\max},
\end{split}
\end{equation}
where each $\tilde v_i$ is a sequence of tokens.

\paragraph{Text-to-latent variable embedding.}
We embed each text span using a pretrained frozen encoder $E$ (token encoder) and a trainable projection:
\begin{equation}
\label{eq:var_embed}
\begin{aligned}
H_i &= E(\tilde v_i)\in\mathbb{R}^{\ell_i\times d_e}, \\
u_i &= \mathrm{pool}(H_i)\in\mathbb{R}^{d_e}, \\
\bar v_i &= \mathrm{Proj}(u_i)\in\mathbb{R}^{d},
\end{aligned}
\end{equation}

\paragraph{Padding and stop mask.}
We pad into fixed slots $V=[v_1,\ldots,v_{N_{\max}}]\in\mathbb{R}^{N_{\max}\times d}$ with a stop mask
$s\in\{0,1\}^{N_{\max}}$:
\begin{equation}
v_i=
\begin{cases}
\bar v_i, & i\le n,\\
\mathbf{0}\in\mathbb{R}^{d}, & i>n,
\end{cases}
\qquad
s_i=
\begin{cases}
1, & i\le n,\\
0, & i>n.
\end{cases}
\end{equation}

\subsection{Conditional masked denoiser (LLaDA-style)}
\label{sec:denoiser}
Let $V^0\in\mathbb{R}^{N_{\max}\times d}$ be the clean variable embeddings.
We learn a mask embedding $m\in\mathbb{R}^{d}$ and mask a subset of \emph{active} slots
$\mathcal{M}\subseteq\{i:s_i=1\}$:
\begin{equation}
V^t_i =
\begin{cases}
m & i\in\mathcal{M},\\
V^0_i & \text{otherwise}.
\end{cases}
\end{equation}
We also encode the full problem prompt into token states $P\in\mathbb{R}^{m\times d}$ and apply cross-attention at each layer. The denoiser $\Theta$ predicts clean embeddings and stop probabilities:
\begin{equation}
(\hat{V}^0,\hat{s})=\Theta(V^t,P),
\end{equation}
where $\hat{V}^0\in\mathbb{R}^{N_{\max}\times d}$ and $\hat{s}\in[0,1]^{N_{\max}}$.

\subsection{Stage I: LLaDA style reconstruction pretraining}
\label{sec:stage1}
We sample a masking rate $t\sim \mathrm{Uniform}(0,1)$ and mask each active slot with probability \ $t$. Reconstruction is applied only to masked active positions:
\begin{equation}
\mathcal{L}_{\text{recon}}
=
\mathbb{E}\Big[
\sum_{i=1}^{N_{\max}}
\mathbf{1}[i\in\mathcal{M}]\cdot
\|\hat{V}^0_i - V^0_i\|_2^2
\Big].
\end{equation}
Stop-mask supervision uses BCE:
\begin{equation}
\begin{split}
    \mathcal{L}_{\text{stop}}&=\mathbb{E}\Big[
\sum_{i=1}^{N_{\max}}
\mathrm{BCE}(\hat{s}_i,s_i)
\Big],\\
\mathcal{L}_{\text{pre}}&=\mathcal{L}_{\text{recon}}+\lambda_{\text{stop}}\mathcal{L}_{\text{stop}}.
\end{split}
\end{equation}
More details can refer to \cite{nie2025large}

\subsection{Stage II: TraceRL-style RL in latent (embedding) space}
\label{sec:stage2}

We finetune the denoiser parameters $\Theta_p$ to directly optimize task reward using a \textbf{trajectory-aware policy optimization} objective~\citep{wang2025revolutionizing}. In this work, both the \emph{reward head} and the \emph{value model} operate on latent variable embeddings rather than the tokens. Therefore the RL loop does \emph{not} require text decoding or rendering.

\paragraph{Latent diffusion trajectory.}
Conditioned on an input $p$, the reverse process induces a latent trajectory $\tau = (V_T, V_{T-1}, \ldots, V_0)$ in continuous space. At each reverse step $t \in \{T,\ldots,1\}$, we sample the next latent state from a Gaussian centered at the denoiser prediction:
\begin{equation}
\label{eq:latent_policy}
\begin{split}
    V_{t-1} \sim \pi_{\Theta_p}(\cdot \mid V_t, p, t) &= \mathcal{N}\!\big(\mu_{\Theta_p}(V_t,p,t), \Sigma_t\big) \\
   \log \pi_{\Theta_p}(V_{t-1}\mid V_t,p,t) &= -\tfrac{1}{2}\|V_{t-1}-\mu_{\Theta_p}(V_t,p,t)\|_{\Sigma_t^{-1}}^2 \\ &+ \text{const},
\end{split}
\end{equation}
where $\mu_{\Theta_p}$ is parameterized by the denoiser and $\Sigma_t$ follows the diffusion schedule.

\paragraph{Reward head in embedding space.}
We compute a scalar reward $r \in [0,1]$ from the \emph{final} latent output $V_0$ using a lightweight scoring rule. For multiple-choice tasks, we score each candidate option by cosine similarity to a predicted \emph{answer embedding} and take an argmax. For exact-match style tasks, we threshold similarity between a predicted \emph{final embedding} and the gold answer embedding. In the verifiable-reward setting, we use terminal-only reward: $r_0=r$ and $r_t=0$ for $t>0$.

\paragraph{Trajectory-aware PPO objective}
We follow TraceRL's \textbf{trajectory-aware PPO} objective with trace shrink. Let $\pi_{\text{old}}$ be a frozen snapshot used to collect rollouts, and define the  ransition density ratio for each step:
\begin{equation}
\label{eq:rho_latent}
\rho_t(\Theta_p)
=
\frac{\pi_{\Theta_p}(V_{t-1}\mid V_t,p,t)}{\pi_{\text{old}}(V_{t-1}\mid V_t,p,t)}.
\end{equation}
We maximize the clipped surrogate with a KL penalty ~\citep{schulman2017proximal}:
\begin{equation}
\label{eq:tracerl_policy_latent}
\begin{aligned}
J_{\text{policy}}(\Theta_p)
&=
\mathbb{E}_{p}\,
\mathbb{E}_{\tau \sim \pi_{\text{old}}(\cdot \mid p)}
\Bigg[
\frac{\sum_{t \in \tau^{(s)}} C_\epsilon\!\big(\rho_t(\Theta_p), A_t\big)}{|\tau^{(s)}|}
\\
&\hspace{3.2em}
-\;
\beta\,\mathrm{KL}\!\left(\pi_{\Theta_p}\,\|\,\pi_{\text{old}}\right)
\Bigg], \\
C_\epsilon(\rho,A)
&=
\min\!\Big(\rho A,\;\mathrm{clip}(\rho,1-\epsilon,1+\epsilon)\,A\Big).
\end{aligned}
\end{equation}
and minimize $\mathcal{L}_{\text{RL}}=-J_{\text{policy}}(\Theta_p)$ as the loss function.
When using terminal-only verifiable rewards, we set $A_t \equiv A$ (shared across the trajectory) by default, matching TraceRL's efficient RLVR regime~\citep{wang2025revolutionizing}.

\paragraph{Latent-space value model and advantages.}
We train a value network $V_{\Theta_v}$ in latent space to reduce variance and match the output of the diffusion model. The value model takes $(V_t,p,t)$ and predicts a step-wise value estimate $V_{\Theta_v,t}$.
We compute advantages with GAE/TD($\lambda$)~\citep{schulman2015high} over reverse steps:
\begin{align}
\delta_t &= r_t + \gamma V_{\text{old},t-1} - V_{\text{old},t}, \\
A_t &= \delta_t + \gamma\lambda A_{t-1},
\end{align}
with terminal conditions consistent with reverse-time indexing.

\paragraph{Clipped value regression.}
We train $V_{\Theta_v}$ with a clipped regression objective following TraceRL:
\begin{equation}
\label{eq:tracerl_value_latent}
\begin{aligned}
J_{\text{value}}(\Theta_v)
&=
\frac{1}{2}\,
\mathbb{E}_{p}\,
\mathbb{E}_{\tau \sim \pi_{\text{old}}(\cdot \mid p)}
\Bigg[
\frac{1}{|\tau^{(s)}|}
\sum_{t \in \tau^{(s)}}
\\
&\hspace{3.2em}
\max\!\Big(
\big(V_{\Theta_v,t} - R_t\big)^2,\; \\
&\hspace{3.2em}\big(V^{\text{clip}}_t - R_t\big)^2
\Big)
\Bigg], \\
V^{\text{clip}}_t
&=
V_{\text{old},t}
+\mathrm{clip}\!\big(V_{\Theta_v,t}-V_{\text{old},t},-\epsilon,\epsilon\big).
\end{aligned}
\end{equation}

where $R_t$ is the return defined by the chosen reward assignment. More details can be found in \cite{}

\paragraph{Overall optimization.}
We alternate between (i) updating $\Theta_p$ via~\eqref{eq:tracerl_policy_latent} and (ii) updating $\Theta_v$
via~\eqref{eq:tracerl_value_latent}. We standardize advantages within a minibatch and tune
$(\epsilon,\beta,\gamma,\lambda)$ for stability, following TraceRL~\citep{wang2025revolutionizing}.

\subsection{Stage III: Rendering with Robust Vec2Text}
\label{sec:render}

We render the final planned variable embeddings into text using \textsc{Vec2Text}~\citep{morris2023text}, an embedding-inversion model based on an iterative hypothesizer-corrector paradigm. The inverse map from a dense embedding to a discrete token sequence is inherently brittle: small perturbations in embedding space can induce discontinuous changes in the decoded text, and planner outputs may drift off the embedder's manifold. Unlike image diffusion, where a continuous decoder maps latents to pixels, text rendering must reconstruct a discrete sequence and is therefore especially sensitive to noise.

\paragraph{Robust Training}
To bridge the distribution gap between training-time embeddings and noisy planner outputs, we robustify the Vec2Text corrector through noise injection. Given a planned variable embedding $v$, we inject $L_2$ noise
\begin{equation}
v' = v + \epsilon,\qquad \epsilon\sim\mathcal{N}(0,\alpha^2 I),
\end{equation}
and train the corrector to recover the exact target answer text from $v'$ (details in Appendix~\ref{app:vec2text}). Concretely, we minimize the standard sequence negative log-likelihood conditioned on the perturbed embedding:
\begin{equation}
\mathcal{L}_{\text{render}}
=
-\mathbb{E}_{(x,v)}\,
\mathbb{E}_{\epsilon\sim\mathcal{N}(0,\alpha^2 I)}
\Big[\log p_{\psi}\!\big(x \mid v', \text{ctx}\big)\Big],
\end{equation}
where ctx denotes the corrector’s auxiliary conditioning (e.g., the current hypothesis and its embedding).
This robust training substantially improves robustness for long variable segments and is a key driver of our
rendering gains over semi-autoregressive alternatives.

\section{Experiments}
\label{sec:experiments}
\subsection{Setup}

\paragraph{Benchmarks.}
We evaluate on nine standard benchmarks spanning general knowledge/reasoning, math/science, and code: MMLU~\citep{hendryckstest2021}, BBH~\citep{suzgun2022challengingbigbenchtaskschainofthought}, ARC-Challenge~\citep{clark-2018-arc}, TruthfulQA~\citep{lin2022truthfulqameasuringmodelsmimic}, PIQA~\citep{bisk2019piqareasoningphysicalcommonsense}
(general); GSM8K~\citep{cobbe2021trainingverifierssolvemath}, MATH~\citep{hendrycks2021measuringmathematicalproblemsolving}, GPQA~\citep{rein2023gpqa} (math/science); and HumanEval~\citep{chen2021evaluatinglargelanguagemodels} (code). We follow LLaDA~\citep{nie2025large} for few-shot prompting.

\paragraph{Baseline.}
Our primary diffusion language model baseline is LLaDA 8B~\citep{nie2025large} under (i) pure diffusion decoding and (ii) the semi-autoregressive / block diffusion sampling variant provided in the same work. We also contextualize results against strong AR models (LLaMA3 8B\cite{chen2021evaluatinglargelanguagemodels}, Qwen2/2.5 7B\cite{yang2024qwen2technicalreport,qwen2025qwen25technicalreport}, Mistral 7B\citep{jiang2023mistral7b}, DeepSeek 7B\citep{jiang2023mistral7b}, Gemma2 9B\citep{gemmateam2024gemma2improvingopen}) using the numbers reported in prior work. As in LLaDA-style reporting, we distinguish \emph{same-protocol} evaluations ($^*$) from sourced results
to avoid overstating strict comparability.

\paragraph{VDLM implementation (planner + renderer).}
VDLM consists of (i) a \emph{variable diffusion planner} (22-layer Transformer denoiser, $d{=}2048$) and (ii) an \emph{Robust-Vec2Text renderer} built on the Vec2Text inversion framework~\citep{morris2023text} with a default LLaMA 3 8B (instruct) embedding encoder. Training proceeds in three stages: \textbf{Stage I} masked diffusion pretraining ; \textbf{Stage II} TraceRL-style RL in embedding space (no decoding inside the RL loop)~\citep{wang2025revolutionizing}; \textbf{Stage III} renderer training with noise perturbations. 
Unless stated otherwise, we run Vec2Text correction for $L' = 10$ iterations at inference.

\subsection{Main Results}
\label{sec:exp_main}
\paragraph{Pre-training.}
Table~\ref{tab:benchmark_main} reports pre-trained results across nine benchmarks. Among the models evaluated under the same protocol ($^*$), VDLM achieves the highest average score (51.6), outperforming LLaDA by 2.3 points (49.3). VDLM improves over LLaDA on MMLU (+5.5), BBH (+4.8), ARC-C (+8.5), and TruthfulQA (+4.1), while matching PIQA. These gains show that diffusion-time refinement in a semantic latent space can preserve the capabilities of token-space diffusion while reducing training and memory costs. 

On mathematical reasoning, VDLM reaches 66.6 on GSM8K and 35.3 on MATH.
Relative to LLaDA, VDLM is slightly lower on GSM8K (-3.7) but higher on MATH (+3.9). On HumanEval, VDLM trails LLaDA (31.9 vs. 35.4), suggesting code generation remains a challenging regime for the current renderer, motivating code-specialized rendering or post-training.

\begin{table*}[t]
\centering
\small
\setlength{\tabcolsep}{5pt}
\renewcommand{\arraystretch}{1.10}
\caption{
\textbf{Benchmark Results (Pre-training).}
VDLM (variable diffusion + Vec2Text) on 9 benchmarks. $^*$ indicates that models are evaluated under the same protocol. The numbers in parentheses represent the number of shots used for in-context learning. ``-'' indicates unknown.
}
\label{tab:benchmark_main}
\begin{tabular}{l|ccc|ccccc}
\toprule
\textbf{Benchmark} & \textbf{VDLM}$^*$ & \textbf{LLaDA}$^*$ & \textbf{LLaMA3} & \textbf{Qwen2} & \textbf{Qwen2.5} & \textbf{Mistral} & \textbf{Deepseek} \\
& \textbf{Diffusion} & \textbf{Diffusion} & \textbf{AR 8B} & \textbf{7B} & \textbf{7B} & \textbf{7B} & \textbf{7B} \\
\midrule
\multicolumn{8}{c}{\textit{General Tasks}} \\
\midrule
MMLU (5) & 71.4 & 65.9  & 65.4 & 70.3 & 74.2 & 64.2 & 48.2 \\
BBH (3) & 54.5 & 49.7  & 62.1 & 62.3 & 70.4 & 56.1 & 39.5 \\
ARC-C (0) & 54.4 & 45.9  & 53.1 & 60.6 & 63.7 & 60.0 & 48.1 \\
TruthfulQA (0) & 50.2 & 46.1  & 44.0 & 54.2 & 56.4 & 42.2 & 57.4 \\
PIQA (0) & 74.2 & 73.6  & 80.6 & — & — & — & 79.2 \\
\midrule
\multicolumn{8}{c}{\textit{Mathematics \& Science}} \\
\midrule
GSM8K (4) & 66.6 & 70.3  & 48.7 & 80.2 & 85.4 & 36.2 & 17.4 \\
MATH (4) & 35.3 & 31.4  & 16.0 & 43.5 & 49.8 & 10.2 & 6.0 \\
GPQA (5) & 25.6 & 25.2  & 25.9 & 30.8 & 36.4 & 24.7 & — \\
\midrule
\multicolumn{8}{c}{\textit{Code}} \\
\midrule
HumanEval (0) & 31.9 & 35.4  & 34.8 & 51.2 & 57.9 & 29.3 & 26.2 \\
\bottomrule
\end{tabular}
\end{table*}

\paragraph{Post-training.}
Table~\ref{tab:posttrain_benchmark} reports post-trained results.VDLM employs SFT followed by TraceRL-style RL in embedding space~\citep{wang2025revolutionizing}, while LLaDA uses SFT only. VDLM improves substantially on generation-heavy tasks: GSM8K increases from 66.6 to 89.8 (+23.2), MATH from 35.3 to 62.4 (+27.1), and HumanEval from 31.9 to 74.9 (+43.0). This confirms that the post-training recipe substantially improves downstream task performance in our setting,  highlighting that embedding-space diffusion post-training is an effective approach.

VDLM narrows the gap with leading AR models. On GSM8K, VDLM (89.8) approaches Qwen2.5- 7B and outperforms LLaMA3-8B Instruct (78.3) by +11.5 points. On MATH, VDLM (62.4) surpasses Qwen2-7B (52.9) by +9.5 points and Gemma2-9B (44.3) by +18.1 points, though it trails Qwen2.5-7B (75.5). Among diffusion models evaluated under the same protocol (*), VDLM outperforms LLaDA by +20.4 on GSM8K and +30.5 on MATH.

Code generation benefits from post-training. VDLM achieves 74.9 on HumanEval, a +43.0 point improvement over pre-trained VDLM (31.9). This result demonstrates that embedding-space RL can effectively optimize for structured output generation, closing the gap with AR models.

\begin{table*}[t]
\centering
\small
\setlength{\tabcolsep}{5pt}
\renewcommand{\arraystretch}{1.10}
\caption{
\textbf{Benchmark Results (Post-training).}
VDLM employs SFT and RL, while LLaDA uses SFT only; other models include RL alignment. $^*$ indicates models evaluated under the same protocol. }
\label{tab:posttrain_benchmark}
\begin{tabular}{ll|ccc|cc|c}
\toprule
& & \multicolumn{3}{c|}{\textit{General Tasks}} & \multicolumn{2}{c|}{\textit{Math \& Science}} & \textit{Code} \\
\textbf{Model} & \textbf{Post-train} & \textbf{MMLU} & \textbf{MMLU-Pro} & \textbf{ARC-C} & \textbf{GSM8K} & \textbf{MATH} & \textbf{HumanEval} \\
& & (5) & (0) & (0) & (4) & (4) & (0) \\
\midrule
\multicolumn{8}{c}{\textit{Diffusion Models}} \\
\midrule
VDLM$^*$ & SFT+RL & 73.9 & 47.8 & 80.8 & 89.8 & 62.4 & 74.9 \\
TraDo 8B$^*$ & SFT+RL & — & — & — & 92.3 & — & — \\
LLaDA 8B$^*$ & SFT & 65.5 & 37.0 & 88.5 & 69.4 & 31.9 & 49.4 \\
\midrule
\multicolumn{8}{c}{\textit{Autoregressive Models}} \\
\midrule
LLaMA3 8B$^*$ & SFT+RL & 68.4 & 41.9 & 82.4 & 78.3 & 29.6 & 59.8 \\
Qwen2 7B$^\dagger$ & SFT+RL & — & 44.1 & — & 85.7 & 52.9 & 79.9 \\
Qwen2.5 7B$^\dagger$ & SFT+RL & — & 56.3 & — & 91.6 & 75.5 & 84.8 \\
Gemma2 9B$^\dagger$ & SFT+RL & — & 52.1 & — & 76.7 & 44.3 & 68.9 \\
Deepseek 7B$^\dagger$ & SFT+RL & 49.4 & — & 49.4 & 63.0 & 15.8 & 48.2 \\
\bottomrule
\end{tabular}
\end{table*}

\subsection{Ablations and Analysis}
\label{sec:exp_ablation}
We conduct ablations on three design choices that are central to VDLM's performance: (1) the encoder backbone architecture, (2) the number of Vec2Text correction iterations, (3) the role of robust training, and (4) the influence of noise level. 

\paragraph{Vec2Text iteration budget.}
Table~\ref{tab:ablation_iterations} varies the number of Vec2Text correction iterations $L'$. Performance improves monotonically from $L'=1$ to $20$, with gains saturating around $L'=20$.

The magnitude of improvement differs substantially by task type. For the short-output task MMLU, increasing iterations from $L'=1$ to $L'=10$ yields a +18.3 point gain. In contrast, long-output tasks exhibit much larger improvements: GSM8K gains +19.1 points, MATH gains +24.2 points, and HumanEval gains +27.5 points. Notably, the marginal returns diminish at higher iteration counts. $L'=20$ approaches the reconstruction ceiling for the current Vec2Text model, and further gains would require improving the renderer itself rather than adding iterations.

This pattern supports the hypothesis that iterative correction primarily mitigates accumulated reconstruction error on long outputs. For multiple-choice tasks like MMLU, a single decoding pass captures most of the signal since outputs are short. However, for multi-step reasoning (GSM8K, MATH) and code generation (HumanEval), where outputs span dozens of tokens, the iterative refinement of Vec2Text becomes critical—each correction step reduces embedding-to-text drift that compounds over longer sequences.

\begin{table*}[t]
\centering
\small
\setlength{\tabcolsep}{7pt}
\renewcommand{\arraystretch}{1.15}
\caption{\textbf{Ablation on number of iteration for Vec2Text.} $L'$ is the number of iteration}
\label{tab:ablation_iterations}
\begin{tabular}{l c |c c  c c}
\toprule
 &  & MMLU & GSM8K & Math & HumanEval \\
\midrule
VDLM
& $L' = 1$  & 55.6 & 70.7 & 38.2 & 47.4 \\
& $L' = 3$  & 68.4 & 80.3 & 57.6 & 64.2 \\
& $L' = 5$ & 72.6& 86.5 & 62.1 & 70.3 \\
& $L' = 10$ & 73.9& 89.8 & 62.4 & 74.9 \\
& $L' = 20$ & 74.1& 89.8 & 62.5 & 75.3 \\
\bottomrule
\end{tabular}
\end{table*}

\paragraph{Renderer robustness}
Table~\ref{tab:ablation_renderer} isolates the impact of robust rendering.
Vanilla Vec2Text (trained only on clean inversions) performs poorly when decoding planner-produced embeddings, while Robust-Vec2Text (trained with $L_2$ perturbations) substantially improves GSM8K/MATH/HumanEval. This supports the central claim that \emph{robust latent-to-text rendering} is a key bottleneck for diffusion-style pipelines when the latent is imperfect at inference.

\begin{table}[t]
\centering
\small
\renewcommand{\arraystretch}{1.15}
\caption{\textbf{Ablation on modules for VLDM}}
\label{tab:ablation_renderer}
\begin{tabular}{l|ccc}
\toprule
 & GSM8K & Math & HumanEval  \\
\hline
Block Diffusion LLaDA  & 77.5 & 42.2 & 46.3 \\
\hline
Vec2Text     & 46.2 & 32.3 & 44.2   \\
Robust-Vec2Text   & 89.8 & 62.4 & 74.9 \\
\bottomrule
\end{tabular}
\end{table}

\paragraph{Embedding encoder backbone.}
Table~\ref{tab:ablation_encoder} shows that VDLM benefits strongly from higher-quality embedding spaces. Using a weak encoder (T5-Base) yields poor results, while stronger encoders markedly improve downstream accuracy. We include OpenAI's \texttt{text-embedding-ada-002} as a widely used embedding baseline~\citep{lin2023vectorsearchopenaiembeddings}.  Further gains emerge from open-weight LLM-based encoders. LLaMA3-8B embeddings achieve 89.8 / 62.4 / 74.9 on GSM8K / MATH / HumanEval, improving over \texttt{ada-002} by +7.1 / +3.5 points on the math benchmarks while slightly trailing on HumanEval (-4.0). The strongest results come from the reasoning model', Qwen3-Embedding-8B. These results show 1) VDLM can leverage future advances in embedding models. 2) the performance gap between T5-Base and Qwen3-Embedding underscores that the semantic structure of the latent space is not incidental but fundamental to enabling effective diffusion-time reasoning.

\begin{table}[t]
\centering
\small
\renewcommand{\arraystretch}{1.15}
\caption{\textbf{Ablation on Encoder Corrector Backbone}}
\label{tab:ablation_encoder}
\begin{tabular}{l|ccccccc}
\toprule
 & GSM8K & Math & HumanEval \\
\hline
T5-Base  &  32.0   & 12.3  & 15.7    \\
\hline
text-embedding-ada-002   & 82.7 & 58.9 & 78.9  \\
\hline
LLAMA3-8B-instruct  & 89.8 & 62.4 & 74.9 \\
\hline
Qwen3-Embedding-8B  & 92.1& 72.0 & 80.4 \\
\bottomrule
\end{tabular}
\end{table}

\paragraph{Noise Level}
Table~\ref{tab:ablation_noise} varies the noise level $\alpha$ used during robust training of Vec2Text. The result shows too little noise leaves the renderer brittle to planner outputs, while excessive noise degrades the learning signal. The optimal noise level is $\alpha = 0.01$. This configuration improves over the lowest noise setting ($\alpha=0.001$) by +34.2 / +14.2 / +25.8 points respectively.

\begin{table}[t]
\centering
\small
\setlength{\tabcolsep}{7pt}
\renewcommand{\arraystretch}{1.15}
\caption{\textbf{Ablation on noise level for Vec2Text.} $\alpha$ is the noise level}
\label{tab:ablation_noise}
\begin{tabular}{l | c c c c c}
\toprule
   & GSM8K & Math & HumanEval \\
\midrule

 $\alpha = 0.001$& 45.6 &48.2 & 49.1 \\
 $\alpha = 0.005$& 78.3 &57.6 & 61.0 \\
 $\alpha = 0.01$ & 89.8& 62.4 & 74.9  \\
 $\alpha = 0.02$ & 80.2& 61.7 & 71.1 \\
 $\alpha = 0.05$ & 76.6& 54.3 & 62.8 \\
\bottomrule
\end{tabular}
\end{table}

\section{Discussion}
\label{sec:discussion}

\paragraph{Renderer as the dominant failure mode.}
Across our ablations, performance is far more sensitive to the rendering backbone than to the diffusion planner itself. Vanilla embedding inversion is brittle under planner-produced noise, and the gap closed by Robust-Vec2Text indicates that \emph{robustness to  embeddings} is a first-order requirement for latent-variable diffusion LMs. This perspective also helps interpret why semi-AR/ block diffusion can degrade when segments grow: local decoding errors become harder to revise as the effective ``block'' of committed content increases, whereas iterative correction in embedding space explicitly revisits the entire hypothesis over multiple rounds.

\paragraph{RL aligns the tasks.}
Stage II optimizes rewards directly on variable embeddings, avoiding rendering inside the RL loop. This design trades token-level supervision for a faster optimization inner-loop, which is especially relevant when the renderer is large and iterative. At the same time, it introduces a new alignment question: the planner is optimized against an embedding-space scoring rule, so the renderer must faithfully map the optimized embeddings back to text. Our results suggest this coupling is workable when the renderer is robustified, but it remains an important axis for future controlled study.

\paragraph{Relation to emerging ``latent reasoning'' directions.}
Recent work has argued that decoupling internal reasoning states from token generation can improve efficiency and/or capability, by letting models do computation in a compact latent space before producing text. Our findings are consistent with this thesis: when the latent space is semantically meaningful and the renderer is strong, latent-space refinement plus decoding can be competitive despite smaller trainable planning capacity. We view VDLM as complementary to these lines of work: rather than introducing an opaque latent, we use explicit variable slots and a separately trained renderer, enabling targeted robustness training and modular encoder upgrades.

\paragraph{Limitations.}
Math/code generation remains challenging in the pre-trained setting, indicating that embedding inversion may not preserve syntactic exactness as reliably as natural language. renderer specialization is a natural next step. Also, post-training improves GSM8K/MATH/HumanEval, further study in the post-training mechanism is required for the future work. 

\section{Conclusion}
\label{sec:conclusion}

We introduced \textbf{VDLM}, a modular variable diffusion language model that separates semantic planning from text rendering. VDLM applies LLaDA-style masked diffusion over semantic variable embeddings, post-trains the planner with TraceRL-style trajectory optimization using \emph{embedding-space} rewards/values, and renders via \textbf{Vec2Text} with \textbf{embedding perturbations} for robustness to planner noise. Across nine benchmarks, VDLM is competitive pre-training and achieves strong post-training gains on long-form generation; ablations show that robust rendering and embedding quality are key drivers. 

\newpage
\bibliography{custom}

\appendix

\appendix
\section{Appendix Overview}
We include implementation details, full preprocessing, extended ablations, and additional robustness
experiments. This appendix is non-essential for the main narrative.

\section{Variable segmentation details}
\label{app:split}
We use a simple segmentation function \texttt{Split} that produces variable chunks from the input text. Concretely, we split by whitespace-delimited line breaks and tab markers, steps, choice and etc, then enforce length bounds and pad to $N_{\max}$.

\section{Reward computation in embedding space}
\label{app:reward}
We compute reward without decoding to text during RL finetuning.
\paragraph{Multiple-choice.}
We embed candidate options with the same frozen encoder and select the argmax cosine similarity.
\paragraph{Exact match.}
We embed the gold answer string and compare against the predicted final embedding via cosine thresholding.
We normalize thresholds on a held-out dev subset.

\section{Vec2Text Rendering Pipeline}
\label{app:vec2text}

This appendix summarizes the \textsc{Vec2Text} embedding-to-text inversion pipeline~\citep{morris2023text}
and specifies the losses used for the hypothesizer--corrector training.

\subsection{Controlled generation view of embedding inversion}
Let $\phi:\mathcal{X}\rightarrow\mathbb{R}^d$ be a frozen text embedder and let $e\in\mathbb{R}^d$ denote a target embedding.
Vec2Text frames inversion as controlled generation, seeking text whose embedding matches $e$:
\begin{equation}
\hat{x} = \arg\max_{x} \cos\!\big(\phi(x), e\big).
\end{equation}

\subsection{Hypothesizer--corrector recursion}
A one-shot inversion model $p(x\mid e)$ is often insufficient, so Vec2Text generates an initial hypothesis and iteratively
refines it. Let $x^{(t)}$ be the hypothesis at iteration $t$ and $\hat{e}^{(t)}=\phi(x^{(t)})$ its embedding. The correction
process can be written as:
\begin{align}
p\!\left(x^{(t+1)} \mid e\right)
&= \sum_{x^{(t)}} p\!\left(x^{(t)} \mid e\right)\,
    p\!\left(x^{(t+1)} \mid e, x^{(t)}, \hat{e}^{(t)}\right), \\
\hat{e}^{(t)} &= \phi\!\left(x^{(t)}\right),
\end{align}
with base distribution (the hypothesizer)
\begin{equation}
p\!\left(x^{(0)} \mid e\right) = p\!\left(x^{(0)} \mid e, \emptyset, \phi(\emptyset)\right).
\end{equation}

\subsection{Training objectives}
\paragraph{Base inversion (hypothesizer).}
Vec2Text trains an initial inversion model by maximum likelihood on paired text/embedding data:
\begin{equation}
\mathcal{L}_{\text{base}} = -\mathbb{E}_{x\sim\mathcal{D}}\left[\log p\!\left(x \mid \phi(x)\right)\right].
\end{equation}

\paragraph{Corrector training.}
Given hypotheses $x^{(t)}$ (seeded from the base model), the corrector is trained with standard sequence NLL:
\begin{equation}
\mathcal{L}_{\text{corr}} =
-\mathbb{E}_{x\sim\mathcal{D}}\,
\mathbb{E}_{x^{(t)}\sim p(\cdot\mid \phi(x))}\left[
\log p\!\left(x \mid \phi(x), x^{(t)}, \phi(x^{(t)})\right)
\right].
\end{equation}

\paragraph{Robust training}
To handle noisy planner embeddings, we perturb the conditioning embedding and train the corrector to recover the
ground-truth text:
\begin{equation}
v' = v + \epsilon,\qquad \epsilon\sim\mathcal{N}(0,\alpha^2 I),
\end{equation}
\begin{equation}
\mathcal{L}_{\text{render}}
=
-\mathbb{E}_{(x,v)}\,
\mathbb{E}_{\epsilon\sim\mathcal{N}(0,\alpha^2 I)}
\Big[\log p_{\psi}\!\big(x \mid v', x^{(t)}, \phi(x^{(t)})\big)\Big].
\end{equation}
In practice we apply this perturbation primarily to the corrector (not the base hypothesizer), since iterative correction
is where robustness to embedding drift matters most for long variable segments.

\section{Implementation Details}
\label{app:implementation}

\subsection{Encoder configuration.}
We use \textbf{LLaMA-3 8B Instruct} as the default encoder.
\begin{itemize}
    \item \textbf{Model}: LLaMA-3 8B Instruct
    \item \textbf{Parameters}: 8B
    \item \textbf{Context length}: 8192 tokens
    \item \textbf{Hidden dimension}: 4096
    \item \textbf{Attention heads}: 32
    \item \textbf{Layers}: 32
    \item \textbf{Vocabulary}: 128{,}256 tokens
\end{itemize}

\subsection{Vec2Text Module}
\label{app:vec2text_module}
For converting between embeddings and natural language (used in hypothesis generation and verification),
we adopt a T5-based default decoder architecture~\citep{raffel2023exploringlimitstransferlearning}.

\subsection{Vec2Text configuration.}
\begin{itemize}
    \item \textbf{Architecture}: T5 encoder-decoder
    \item \textbf{Model size}: T5-large
    \item \textbf{Decoder layers}: 22
    \item \textbf{Hidden dimension}: 2048
    \item \textbf{Feed-forward dimension}: 4096
    \item \textbf{Attention heads}: 16
\end{itemize}

\subsection{VDLM configuration.}
\begin{itemize}
    \item \textbf{Layers}: 22
    \item \textbf{Model dimension}: 2048
    \item \textbf{Attention heads}: 32
    \item \textbf{Feed-forward dimension}: 4096
    \item \textbf{Key/value heads}: 4 (grouped-query attention)
\end{itemize}

\section{AI Assistant Usage}
This research utilized AI assistants including Claude and GPT-5 for several aspects of the paper and dataset preparation. We employed these tools mainly for:
\begin{itemize}
    \item Implementation support: AI assistants provided code debugging assistance for the implementation and modification of dLLM, vec2text repos.
    \item Manuscript preparation: We used AI assistants for literature review to identify relevant papers, proofreading, language refinement, and formatting assistance.
    \item Benchmark: We use AI assistant GPT-5 to source some results from \cite{nie2025large,wang2025revolutionizing}.
\end{itemize}
\end{document}